\documentclass[runningheads]{llncs}
\usepackage{url}
\usepackage{xcolor}
\usepackage{hyperref}
\usepackage{lineno}
\usepackage{float}
\usepackage{amsmath,amsfonts}
\usepackage{algorithmic}
\usepackage{graphicx}
\usepackage{subfigure}
\usepackage{booktabs}
\usepackage{array}
\usepackage{multirow}
\usepackage{bbding}
\usepackage{graphbox}
\usepackage{color}
\usepackage{amssymb}
\usepackage{latexsym}
\usepackage{bbm}
\usepackage{newtxtext,newtxmath}
\usepackage{framed,multirow}
\usepackage[hang,flushmargin]{footmisc}
\usepackage[para,online,flushleft]{threeparttable}
\usepackage{pifont}
\newcommand{\cmark}{\ding{51}}%
\newcommand{\xmark}{\ding{55}}%

\usepackage{graphicx}

\begin{document}
%
%
\title{K-Diag: Knowledge-enhanced Disease Diagnosis in \\ Radiographic Imaging}
%
%
\author{Anonymous Paper ID: 808}
\author{
Chaoyi Wu \inst{1,2,*}  \and
Xiaoman Zhang \inst{1,2,*}  \and
Yanfeng Wang \inst{1,2} \and \\[3pt]
Ya Zhang \inst{1,2} \and
Weidi Xie \inst{1,2,\dag}}

\authorrunning{C. Wu et al.}
\titlerunning{K-Diag: Knowledge-enhanced Disease Diagnosis in Radiographic Imaging}
%
\institute{Cooperative Medianet Innovation Center, Shanghai Jiao Tong University, Shanghai, China \\
\and Shanghai AI Laboratory, Shanghai, China\\
\url{https://chaoyi-wu.github.io/K-Diag/}
}
%

\maketitle              

\vspace{-0.5cm}
\begin{abstract}
In this paper, we consider the problem of disease diagnosis.
Unlike the conventional learning paradigm that treats labels independently, 
we propose a knowledge-enhanced framework, 
that enables training visual representation with the guidance of medical domain knowledge. In particular, we make the following contributions:
{\bf First}, to explicitly incorporate experts' knowledge,
we propose to learn a neural representation of medical knowledge graph via contrastive learning, implicitly establishing relations between different medical concepts. {\bf Second}, while training the visual encoder, we keep the parameters of
the knowledge encoder frozen and propose to learn a set of prompt vectors for efficient adaptation.
{\bf Third}, we adopt a Transformer-based disease-query module for cross-model fusion,
which naturally enables explainable diagnosis results via cross attention.
To validate the effectiveness of our proposed framework, 
we conduct thorough experiments on three x-ray imaging datasets across different anatomy structures, showing our model can exploit the implicit relations between diseases/findings, thus is beneficial to the commonly encountered problem in the medical domain, namely, long-tailed and zero-shot recognition, which conventional methods either struggle or completely fail to realize.


\end{abstract}
\renewcommand{\thefootnote}{}
\footnotetext{*: These authors contribute equally to this work.}
\footnotetext{\dag: Corresponding author.}

\section{Introduction}

\begin{figure}[t]
    \centering
    \includegraphics[width=\linewidth]{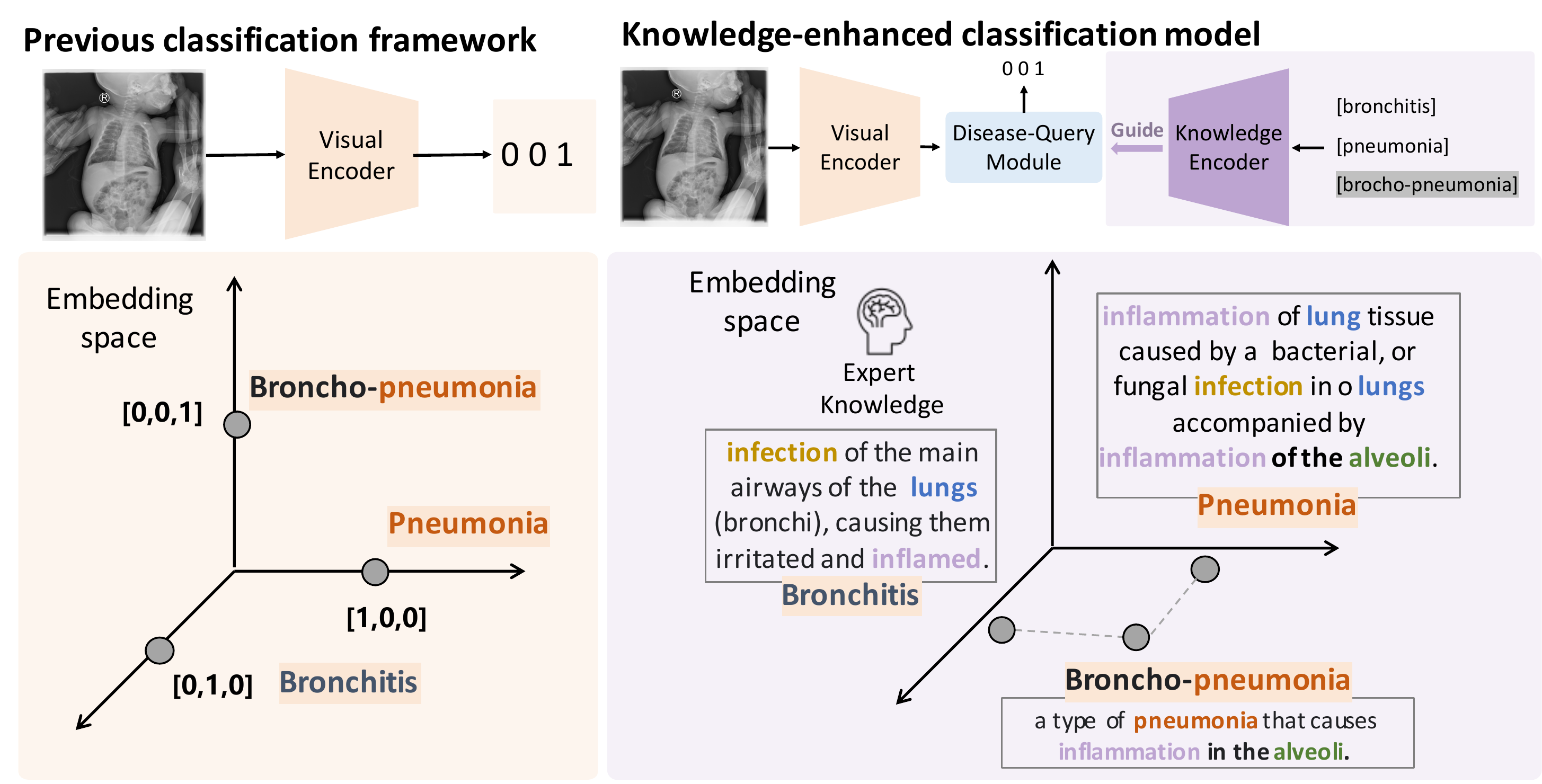}
    \vspace{-0.5cm}
    \caption{In the conventional training scheme~(left), 
    manual annotations are often converted into \textbf{discrete} one-hot vectors, 
    that are \textbf{orthogonal} in the embedding space, thus ignoring the implicit relations between labels. While in our proposed knowledge-enhanced classification framework (right), the labels are transformed into \textbf{continuous} vectors in a knowledge embedding space, to capture the implicit relations, and further used to supervise visual representation learning.
    } 
    \label{fig:intuition}
    \vspace{-0.5cm}
\end{figure}

The application of artificial intelligence (AI) has delivered impressive results in diagnosing diseases from medical scans~\cite{Mirbabaie2021}. 
A commonly adopted framework is to train vision models by supervised learning with discrete labels and predict pathology categories within a fixed-size vocabulary at inference time~\cite{cohen2022torchxrayvision,Zhou2022longtail}.
However, such a learning paradigm suffers from two limitations: 
{\em first}, the model is unable to generalize toward previously unseen categories;
{\em second}, the labels are converted into one-hot vectors as illustrated in Fig.~\ref{fig:intuition}, that are \textbf{orthogonal} in the embedding space, 
leaving the intrinsic relations between different pathologies or diseases unexploited.

In the recent literature, 
jointly training visual-language models has shown promising progress in computer vision~\cite{pratt2022does,radford2021learning}, often called Foundation Models. 
For example, CLIP~\cite{radford2021learning} and ALIGN~\cite{jia2021scaling} have 
demonstrated remarkable ``zero-shot'' generalization for various downstream tasks by learning the joint representation of image and text with simple noise contrastive learning. Crucially, the data used to train these powerful foundation models can simply be crawled from the Internet without laborious manual annotation.
However, as commonly known, 
collecting training data at scale in the medical domain is often impractical~\cite{cohen2022torchxrayvision,bakator2018deep}, 
due to its safety-critical nature.
In this paper, we, therefore, explore an alternative by injecting medical expert knowledge into the visual representation learning procedure.

We present a novel knowledge-enhanced classification framework, as shown in the upper right of Fig.~\ref{fig:intuition}, {\em first}, to explicitly incorporate experts' knowledge,
we build a neural representation for the medical knowledge graph via contrastive learning, acting as a ``knowledge encoder'' that explicitly encodes the relations between medical concepts in the embedding space; 
{\em second}, while training the visual encoder, we keep the parameters of the knowledge encoder frozen, and only learn a set of prompt vectors for efficient adaptation;
{\em third}, we adopt a Transformer-based disease-query module for text-image cross-modality fusion, where the disease names act as queries that cross-attend the visual features and infer the likelihood of the disease existence,
naturally, it enables explainable diagnosis results via the
cross attentions.

To demonstrate the effectiveness of the proposed knowledge-enhanced classification framework, we conduct thorough experiments to analyze from three perspectives:
{\em first}, we experiment on three disease diagnosis tasks across different anatomy structures, to show our method is efficient across various image distributions;
{\em second}, we train on a combination of 11 public chest x-ray datasets, 
showing our model can better exploit the potential of various public datasets regardless of their annotation granularity, 
which traditional training paradigm can slightly benefit from;
{\em third}, we perform zero-shot disease diagnosis, 
{\em i.e.}, evaluating unseen classes on PadChest~\cite{bustos2020padchest}, achieving an AUC of at least 0.600 on 79 out of 106 unseen radiographic findings, 
Note that, such a task is completely unachievable in conventional supervised training.

\section{Method}
In this section, we start by describing the considered problem scenario in Sec.~\ref{sec:problem_scenario}, 
followed by the procedure of condensing the medical domain knowledge into a text encoder in Sec.~\ref{sec:text_encoder}.
In Sec.~\ref{sec:training}, we detail the proposed knowledge-enhanced classification model, including the visual encoder, knowledge encoder, 
learnable prompt module, and disease-query module, and describe the training procedure.
%

\subsection{Problem Scenario}
\label{sec:problem_scenario}
Given a dataset with $N$ sample pairs, 
{\em i.e.}, $\mathcal{D}_{\text{train}} = \{(x_1, y_1), (x_2, y_2),\dots, (x_N, y_N)\}$,
where ${x}_i \in \mathbb{R}^{H \times W \times 3}$ refers to the input image, 
and $y_i \in \mathcal{T}=\{t_1. \dots, t_Q \}$ denotes the ground truth annotation from a pool of $|Q|$ candidate diseases. Unlike conventional supervised learning that often converts the labels to one-hot vectors,
our goal is to train a classification model that leverages the semantics encapsulated in the disease category texts, specifically, 
\begin{align}
    {S}_i = \Phi_{\text{query}}(\Phi_{\text{visual}}({x}_i), \Phi_{\text{prompt}}(\Phi_{\text{knowledge}}(\mathcal{T}))),
\end{align}
where ${S}_i \in \mathbb{R}^{Q}$ refers to the inferred likelihood of the patient having any disease in $\mathcal{T}$.
$\Phi_{\text{knowledge}}(\cdot), \Phi_{\text{prompt}}(\cdot), \Phi_{\text{visual}}(\cdot),  \Phi_{\text{query}}(\cdot)$ refer to the trainable modules in our proposed knowledge-enhanced classification framework, that will be detailed in the following sections.

\subsection{Knowledge Encoder}
\label{sec:text_encoder}
\begin{figure}[!t]
    \centering
    \includegraphics[width=\linewidth]{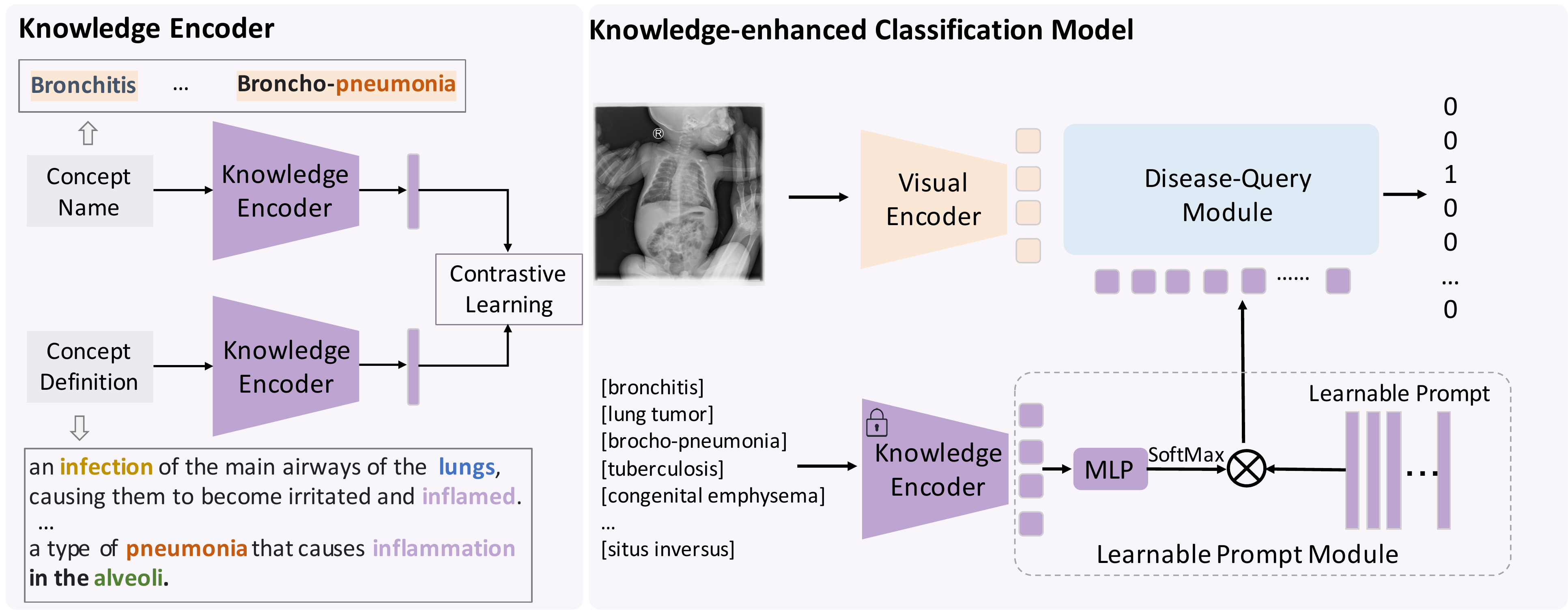}
    \vspace{-0.6cm}
    \caption{Overview of the knowledge-enhanced disease diagnosis workflow. 
    The knowledge encoder~(left) is first trained to learn a neural representation of the medical knowledge graph via contrastive learning, 
    and then used to guide the visual representation learning in our knowledge-enhanced classification model~(right).
    }
    \label{fig:framework}
    \vspace{-0.3cm}
\end{figure}

To explicitly incorporate experts’ knowledge, 
we propose to inject the medical domain knowledge into a text encoder~($\Phi_{\text{knowledge}}$), by implicitly modeling the relations between medical entities in the textural embedding space.
Specifically, we employ an off-the-shelf knowledge graph in the medical community, 
namely, Unified Medical Language System~(UMLS)~\cite{bodenreider2004unified},
to fine-tune a pre-trained BERT language model. 
In the following section, we detail the training procedure, as shown in Fig.~\ref{fig:framework}.

\vspace{6pt} \noindent \textbf{Notation. }
Let $\mathcal{D}_{\text{UMLS}} = \{(n,d)_i\}_{i=1}^{||D||}$ denote a concept dictionary for UMLS in text form, 
each concept~($n_i$) is associated with one corresponding definition~($d_i$),
for example, the concept ``pulmonary infiltrate'' is defined as ``A finding indicating the presence of an inflammatory or neoplastic cellular infiltrate in the lung parenchyma''.

\vspace{6pt} \noindent \textbf{Training.~}
Here, we train the text encoder by maximizing the similarities between positive concept-definition pairs, 
{\em i.e.}, pull the distance of language description and its corresponding concept in the textual embedding space. Given $N$ randomly sampled concepts and definitions, 
we pass them through a standard BERT architecture~\cite{devlin2018bert},
and take the average-pooled features as their textual embeddings.
{\em i.e.}, the concept~$\boldsymbol{n}_i \in \mathbb{R}^{N \times d}$, definition~$\boldsymbol{d}_i\in \mathbb{R}^{N \times d}$. 
At training time,
each mini-batch can be expressed as $\{(n_i,d_i)\}_{i=1}^{N}$,
and the model can be trained via contrastive learning~\cite{oord2018representation}:
\begin{align}
\mathcal{L}_{\text{contrastive}} =  -\frac{1}{2N}\sum_{k=1}^{N}(\log \frac{e^{(\langle{\boldsymbol{n}}_i, {\boldsymbol{d}}_i\rangle/\tau)}}{\sum_{k=1}^{N} e^{(\langle{\boldsymbol{n}}_i, {\boldsymbol{d}}_k\rangle/\tau)}} + \log \frac{e^{(\langle{\boldsymbol{d}}_i, {\boldsymbol{n}}_i\rangle/\tau)}}{\sum_{k=1}^{N} e^{(\langle{\boldsymbol{d}}_i, {\boldsymbol{n}}_k\rangle/\tau)}}).
\end{align}
where $\tau \in \mathbbm{R}^{+}$ is a scalar temperature parameter.
Once this is trained, the text encoder effectively becomes a ``knowledge encoder'' with domain experts' knowledge injected.

\subsection{Knowledge-enhanced Classification Model}
\label{sec:training}

After injecting the domain knowledge into the text encoder, 
here, we describe the procedure to guide the visual representation learning with the knowledge encoder. 
Specifically, the classification model consists of four core modules, 
namely, the visual encoder, frozen knowledge encoder,
prompt module, and disease-query module.

\vspace{6pt} \noindent \textbf{Visual Encoder. }
Given an image scan $x_i \in \mathbb{R}^{H \times W \times 3}$,
we compute the features with a visual backbone, 
which can be either ResNet~\cite{he2016deep} or Vision Transformer~\cite{dosovitskiy2020vit},
$ \boldsymbol{x}_i = \Phi_{\text{visual}}({x}_i) \in \mathbbm{R}^{h\times w \times d}$,
where $d$ refers to the feature dimension,
and $h, w$ denote the size of the output feature map,
feature dimension $d$ is set to 256.


\vspace{6pt} \noindent \textbf{Frozen Knowledge Encoder. }
Given the disease categories $\mathcal{T}$, 
we compute the features with the pre-trained knowledge encoder:
$\boldsymbol{T} = \Phi_{\text{knowledge}}(\mathcal{T}) \in \mathbbm{R}^{Q \times d}$,
where $d$ refers to the feature dimension, and $Q$ refers to the category number.
As the number of classes in downstream medical diagnosis datasets is usually extremely limited, \emph{e.g.}, 10 diseases on VinDr-Mammo~\cite{Nguyen2022VinDrMammoAL}, 
to prevent the knowledge encoder from over-fitting on certain training classes, 
deviating from the originally embedded knowledge graph,
we keep its parameters frozen, and use it to guide the learning of visual encoder, effectively, such a training procedure resembles knowledge injection into visual representation learning.



\vspace{6pt} \noindent \textbf{Learnable Prompt Module. }
To bring more flexibility, we also introduce a learnable prompt module~($\Phi_{\text{prompt}}$) for efficient knowledge adaptation.
Specifically, as shown in the lower-right of Fig.~\ref{fig:framework},
it consists of a set of learnable vectors, {\em i.e.}, $\boldsymbol{h}\in \mathbbm{R}^{N \times d}$, where $N$ denotes the feature numbers and $d$ is the embedding dimension of each feature. 
Given the disease embeddings~($\boldsymbol{T} \in \mathbbm{R}^{Q \times d}$), 
we first use an MLP to project it into a probability distribution over the learnable prompt vectors, $\boldsymbol{p} = \text{SoftMax}(\text{MLP}(\boldsymbol{T})) \in \mathbbm{R}^{Q \times N}$.
Then, the output of the Prompt Module can be calculated as the matrix multiplication between $\boldsymbol{p}$ and $\boldsymbol{h}$, 
{\em i.e.}, that can be formulated as 
$\boldsymbol{k} = \Phi_{\text{prompt}}(T) = (\boldsymbol{p} \cdot \boldsymbol{h}) \in  \mathbbm{R}^{Q \times d}$.


\vspace{6pt} \noindent \textbf{Disease-Query Module. }
We use a 4-layer transformer decoder with an MLP to get the final prediction. 
Given the disease categories $\mathcal{T}$,
we have converted them into a set of disease embeddings~($\boldsymbol{k}$) with the pre-trained knowledge encoder and learnable prompt module.
As inputs to the Transformer decoders,
disease embeddings are treated as \texttt{queries},
and the encoded features~($\boldsymbol{x}_i$) act as \texttt{key} and \texttt{value} of the disease-query module:
${s}_i = \Phi_{\text{query}}(\boldsymbol{x}_i, \boldsymbol{k}) \in \mathbbm{R}^{Q \times C}$, 
where $C$ represents the class number and is set as 2, 
since the diagnosis tasks are all binary classification.
Cross-entropy loss is used as the optimization function.

\section{Experiments}

\subsection{Datasets}
In this paper, we conduct experiments on datasets of X-ray images, 
due to there exists sufficient data across anatomy structures and annotated pathology categories in this field, supporting thorough evaluation.


\vspace{5pt} \noindent \textbf{VinDr-PCXR~\cite{Nguyen2022VinDrPCXRAO}}
is a new pediatric CXR dataset of 9,125 studies, which was officially divided into a training set and a test set of 7,728 and 1,397 studies respectively. Each scan in the training set was manually annotated for the presence of 15 diseases by a pediatric radiologist who has more than ten years of experience,
while in the official test set, there are 11 diseases.
Additionally, for fair and robust evaluation, 
we further merged the rare diseases~(positive samples less than 5) into ``other diseases'',
resulting in only 6 classes in the test set, including no finding, bronchitis, broncho-pneumonia, other disease, bronchiolitis, and pneumonia.


\vspace{5pt} \noindent \textbf{VinDr-Mammo~\cite{Nguyen2022VinDrMammoAL}}
is a full-field digital mammography dataset comprising 20,000 images~(5,000 four-view scans). Each scan was manually annotated for no finding or the presence of 10 mammography findings, including mass, calcification, asymmetry, focal asymmetry, global asymmetry, architectural distortion, and suspicious lymph node, skin thickening, skin retraction, nipple retraction. The dataset was official divided into a training set and a test set with 4,000 and 1,000 exams respectively.

\vspace{5pt} \noindent \textbf{VinDr-SpineXr~\cite{Nguyen2021VinDrSpineXRAD}}
is a spine X-ray dataset comprising 10,468 spine X-ray images from 5,000 studies. 
Each image was manually annotated by an experienced radiologist with no finding or abnormal findings in 7 categories, including osteophytes, foraminal stenosis, vertebral collapse, disc space narrowing, spondylolysthesis, surgical implant, and other lesions. 
The dataset was official divided into a training set a test set of 4,000 and 1,000 studies respectively.

\vspace{5pt} \noindent \textbf{PadChest~\cite{bustos2020padchest}} is a chest X-ray dataset with 160,868 chest X-ray images labeled with 174 different radiographic findings, 19 differential diagnoses, only $27\%$ of the labels~(totaling 39,053 examples) come from board-certified radiologists, and the rest are obtained by using a recurrent neural network with attention trained on the radiology reports. 
For evaluation purposes, we only test on samples annotated by board-certified radiologists, and report the zero-shot test results.

\vspace{5pt} \noindent \textbf{CXR-Mix.}
In this paper, we also construct a dataset by assembling 11 public datasets, 
including {ChestXray-14~\cite{wang2017chestx}}
{GoogleNIH~\cite{majkowska2020chest}}
{Covid-CXR2~\cite{pavlova2021covid}}
{CheXpert~\cite{irvin2019chexpert}}
{Object-CXR~\cite{objectcxr}}
{NLM-TB~\cite{jaeger2014two}}
{RSNA~\cite{shih2019augmenting}}
{SIIM-ACR~\cite{filice2020crowdsourcing}}
{VinDR~\cite{nguyen2022vindr}}
{OpenI~\cite{demner2016preparing}}
{MIMIC-CXR~\cite{johnson2019mimic}},
termed as \textbf{CXR-Mix}.
We refer the readers to~\cite{cohen2022torchxrayvision} for more details of these datasets. For the datasets~\cite{wang2017chestx,irvin2019chexpert} with official train/val/test splits, we use them directly, 
for those~\cite{majkowska2020chest,nguyen2022vindr,pavlova2021covid,objectcxr} with official train/test splits, 
we random split the train split with $0.8/0.2$ for train/val, in other cases, 
we randomly split the datasets~\cite{shih2019augmenting,filice2020crowdsourcing,jaeger2014two,demner2016preparing,johnson2019mimic} with $0.7/0.1/0.2$ for train/val/test. 
As a result, our constructed CXR-Mix ends up with $763,520$ chest X-rays for training, 
$28, 925$ for val, and $28,448$ for test, spanning across a total of $38$ classes in the CXR-Mix. 
\textbf{Note that}, each of these datasets was originally collected to serve different purposes, 
the annotations are often partially available,  
for example, the images from the  pneumonia dataset lack labels for pneumothorax, 
we use $-1$ to denote the label missing and will not calculate the final CE loss on them.


\subsection{Implementation and Training Details}

\vspace{5pt} \noindent \textbf{Knowledge-Enhanced Text Encoder.}
To construct the knowledge encoder, we initialize the text encoder from ClinicalBERT~\cite{alsentzer2019publicly}, and finetune it for 100K training steps.
In each mini-batch, 64 concept-definition pairs are used for training.
We set the maximal sequence length as 256, though the definition could be long sometimes.
We use AdamW~\cite{loshchilov2017decoupled} as the optimizer with $lr =1\times 10^{-4}$ and $lr_\text{warm up} = 1\times 10^{-5}$. 

\vspace{5pt} \noindent \textbf{Knowledge-Enhanced Classification Framework.}
We freeze the knowledge encoder and set other parts to be learnable, 
{\em i.e.}, visual encoder, prompt module, disease-query module.
We train the model for 100 epochs with batch size 128, 
and use AdamW~\cite{loshchilov2017decoupled} is adopted as the optimizer with $lr =1\times 10^{-4}$ and $lr_\text{warm up} = 1\times 10^{-5}$. 

\section{Results}
Here, we conduct experiments to validate the effectiveness of our knowledge-enhanced classification model.
In Sec.~\ref{sec:result1}, we first compare knowledge-enhanced training with standard training using discrete labels across different architectures and then perform a thorough ablation study of proposed modules.
In Sec.~\ref{sec:result2}, we conduct an analysis on the proposed knowledge encoder by replacing it with other pre-trained language models. In Sec.~\ref{sec:result3}, we experiment on \textbf{CXR-Mix}, 
to show that our model can effectively exploit knowledge across datasets with varying annotation granularity, we conduct analysis from three aspects: 
(i) the ability to combine various partial-labeled datasets, 
(ii) to leverage implicit relations between diseases, 
(iii) to diagnose diseases that are unseen at training time, 
resembling an open-set recognition scenario.  



\vspace{-0.2cm}
\subsection{Analysis of Knowledge-Enhanced Classification Model}
\label{sec:result1}

\subsubsection{Comparison to Conventional Training Scheme.}
As baselines, we adopt the widely used ResNet-50~\cite{he2016deep} and ViT-16~\cite{dosovitskiy2020image},
and train with conventional learning scheme, {\em i.e.}, using discrete labels.
The results are summarized in Tab.~\ref{resnet}, 
we refer the reader to detailed results for each category presented in the supplementary material~(Tab.~\ref{pcxr},~\ref{spinexr} and ~\ref{mammo}). 
Our proposed knowledge-enhanced model achieves a higher average AUC on all three datasets across different architectures.

\begin{table}[t]
\caption{Compare with Baseline Models with ResNet-50~\cite{he2016deep} and ViT-16~\cite{dosovitskiy2020vit}  as backbone on disease classification tasks.
KE indicates the proposed knowledge encoder, LP indicates the proposed learnable prompt module, and the number denotes the prompt number.
AUC scores averaged across different diseases are reported. We report the mean and standard deviation of three different seeds.}
\vspace{-5pt}
\centering
\setlength{\tabcolsep}{5pt}
\footnotesize
\begin{tabular}{llcc|ccc}
\toprule
Model & Backbone & KE & LP & VinDr-PCXR & VinDr-Mammo & VinDr-SpineXr  \\
\midrule
Res~\cite{he2016deep} & ResNet-50 & \xmark  & \xmark & 70.53 $\pm$ 1.00 &  82.54 $\pm$ 1.79  & 87.35 $\pm$ 0.38 \\
Res+KE & ResNet-50 & \cmark  &  \xmark & 71.39 $\pm$ 0.66 & 83.23 $\pm$ 1.98 & 87.76 $\pm$ 0.41  \\
Res+KE+LP & ResNet-50 & \cmark & 32 & \textbf{73.97 $\pm$ 0.28} &  80.59 $\pm$ 3.11 & 88.46 $\pm$ 0.22 \\
Res+KE+LP & ResNet-50  & \cmark  & 64 & 72.88 $\pm$ 0.19 &  81.27$\pm$ 1.83  & \textbf{88.90 $\pm$ 0.04 } \\
Res+KE+LP & ResNet-50 & \cmark  & 128 & 73.70 $\pm$ 0.41 & \textbf{ 84.80 $\pm$ 1.04 } & 88.22 $\pm$ 0.53  \\
\midrule
ViT~\cite{dosovitskiy2020vit} & ViT-16 & \xmark   & \xmark & 69.06$\pm$ 0.74 & 80.50 $\pm$ 2.47 & 85.56 $\pm$ 0.97  \\
ViT+KE & ViT-16 & \cmark   & \xmark  & 71.69 $\pm$ 1.64 & 83.89 $\pm$ 0.38  & 85.75 $\pm$ 0.44  \\
ViT+KE+LP & ViT-16 & \cmark  & 32 & \textbf{72.90 $\pm$ 0.97} & 83.67 $\pm$ 2.34 & 86.55 $\pm$ 0.48 \\
ViT+KE+LP & ViT-16 & \cmark  & 64 &  71.07 $\pm$ 3.28 & 84.33 $\pm$ 1.54 & \textbf{86.83 $\pm$ 0.81} \\
ViT+KE+LP & ViT-16 & \cmark  & 128 & 72.47 $\pm$ 0.78 & \textbf{84.46 $\pm$ 0.84 }& 86.33 $\pm$ 0.23  \\
\bottomrule
\end{tabular}
\label{resnet}
\vspace{-0.4cm}
\end{table}

\vspace{-0.3cm}
\subsubsection{Ablation Study.}
We conduct a thorough ablation study of the proposed model by removing individual modules and varying the hyper-parameters, as shown in Tab.~\ref{resnet}.
Specifically, the performance of ResNet-50 is improved to $71.39\%$, $83.23\%$, and $87.76\%$ for the three different tasks equipped with the proposed knowledge encoder to guide the visual representation learning.
While combining with the learnable prompt~(LP) module, which potentially offers more flexibility for knowledge adaptation, we observe a significant performance gain, up to $2.58\%$ on average AUC scores on VinDr-PCXR.
The same conclusion can be drawn for the ViT-16 visual backbone.

To analyze the effect of prompts, we experiment with different numbers of prompts.
As shown, the optimal number of prompts varies from task to task, but in general, adding the LP module benefits all the downstream tasks.
The only exception is for the VinDR-Mammo task with ResNet as a backbone,
which is mainly caused by the extremely small test samples in some categories~(Tab.~\ref{mammo}), {\em e.g.}, skin retraction and skin thickening.

\vspace{-0.3cm}
\subsubsection{Qualitative Visualisation.}
\begin{figure}[!htb]
    \centering
    \includegraphics[width=.97\linewidth]{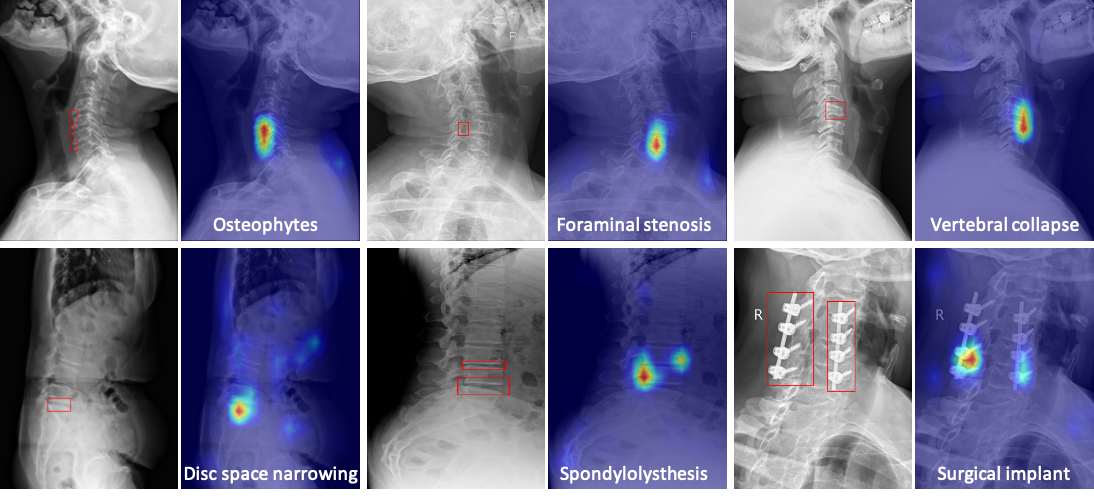}
    \vspace{-0.4cm}
    \caption{Sample visualization of randomly chosen samples from VinDr-SpinXr, 
    we present both the original image (left) and an attention map generated from our proposed model with ResNet-50 as the backbone (right). }
    \label{fig:visualize}
    \vspace{-0.3cm}
\end{figure}

To provide a visualization that can potentially be used for clinicians to discover and understand the evidence that AI algorithm bases its predictions on,
the disease-query module in our proposed architecture enables detailed visualization for each of the queries with positive output.
Specifically, we average the cross-attention map in each transformer layer in the disease query module, and visualize the results in Fig.~\ref{fig:visualize}.
The model's attention well matches radiologists’ diagnoses of different diseases, 
{\em i.e.} red boxes labeled by board-certified radiologists.

\subsection{Analysis of the Knowledge-Enhanced Text Encoder}
\label{sec:result2}
Here we investigate another way of incorporating prior knowledge,
that is, to guide the visual representation with a text encoder pre-trained on the large medical corpus,
such as the electronic health records MIMIC III~\cite{Johnson2016MIMICIIIAF} or scientific publications PubMed, to guide the classification task.
As shown in Tab.~\ref{ablation},
while comparing with the models that adopts ClinicalBERT~\cite{alsentzer2019publicly} or PubMedBERT~\cite{gu2021domain} as knowledge encoder, we can make two observations:
(i) guiding visual representation learning with domain knowledge generally works better,
{\em e.g.}, results of using ClinicalBERT or PubMedBERT outperform conventional training with discrete labels, 
(ii) our proposed knowledge-enhanced text encoder consistently demonstrates superior results, that can be attributed to the explicitly injected domain knowledge, rather than implicitly learning it from the document corpus.


\vspace{-0.2cm}
\begin{table}[!htb]
\caption{Ablation study on knowledge encoder with ResNet as a backbone, 
mean and standard deviation for AUC scores is reported with three different seeds.
we use the optimal prompt numbers according to the ablation study, 
{\em i.e.}, 32 for VinDr-PCXR, 128 for VinDr-Mammo, and 64 for VinDr-SpineXr.}
\vspace{-0.2cm}
\centering
\setlength{\tabcolsep}{5pt}
\renewcommand{\arraystretch}{1.2}
\begin{tabular}{lc|ccc}
\toprule
Model & Knowledge Encoder  & VinDr-PCXR & VinDr-Mammo & VinDr-SpineXr \\
\midrule
Res~\cite{he2016deep} & - & 70.53 $\pm$ 1.00 &  82.54 $\pm$ 1.79  & 87.35 $\pm$ 0.38 \\
Res+KE+LP & ClinicalBERT~\cite{alsentzer2019publicly}  & 70.83 $\pm$ 0.96 & 83.55 $\pm$ 0.72 & 88.19 $\pm$ 0.25\\
Res+KE+LP & PubMedBERT~\cite{gu2021domain}& 71.73 $\pm$ 0.75  & 81.96 $\pm$ 1.82 & 88.25 $\pm$ 0.65 \\
Res+KE+LP & Ours & \textbf{73.97 $\pm$ 0.28} & \textbf{84.80 $\pm$ 1.04} & \textbf{88.90 $\pm$ 0.04 }\\
\bottomrule
\end{tabular}
\label{ablation}
\vspace{-0.4cm}
\end{table}

\subsection{Analysis on the CXR-Mix}
\label{sec:result3}

In this section, we experiment on the assembled dataset \textbf{CXR-Mix}, 
to demonstrate the effectiveness of our proposed framework in exploiting knowledge across datasets and annotation granularity. 

\subsubsection{The Ability to Combine Various Partial-labeled Datasets: }
Unlike the traditional approach, which requires to carefully merging the label space from different datasets~\cite{cohen2022torchxrayvision,cohen2020limits,lambert2020mseg} to benefit from them, 
our formulation of embedding the `disease name' with a knowledge encoder naturally enables us to train models on the mixture of multiple datasets, handling different granularities of diagnosis targets and inconsistent pathology expression. 
As shown in Tab.~\ref{Assembling results}, we compare to TorchXRayVision~\cite{cohen2022torchxrayvision} that merges the label space 
and trains a baseline model with discrete labels, 
our knowledge-enhanced framework improves the performance from $82.60\%$ to $85.13\%$ and form $77.39\%$ to $79.54\%$ under the two commonly-used backbones, ResNet and ViT, respectively. 

\vspace{-0.4cm}
\begin{table}[!htb]
\centering
\caption{Compare with Baseline Models on disease classification tasks on the assembling dataset. AvgAUC refers to the AUC score averaged across different diseases. The first line refers to the use of the training flow proposed by TorchXrayVision~\cite{cohen2022torchxrayvision} and use ResNet or ViT as the backbone.}
\setlength{\tabcolsep}{5pt}
\renewcommand{\arraystretch}{1.2}
\begin{tabular}{lcc|lcc}
\toprule
Methods & Prompt &AvgAUC  & Methods & Prompt & AvgAUC \\ \toprule

Res~\cite{cohen2022torchxrayvision}      & -        & 82.60 $\pm$ 1.27  & ViT~\cite{cohen2022torchxrayvision} & - & 77.39 $\pm$ 0.71\\
Res+KE      & -   & 83.11 $\pm$ 0.27   & ViT+KE & - & 78.30 $\pm$ 0.93\\
Res+KE+LP     & 32   &   84.45 $\pm$ 1.06 & ViT+KE+LP  & 32 & 79.25 $\pm$ 1.05\\
Res+KE+LP     & 64   & \textbf{85.13} $\pm$  0.78 & ViT+KE+LP   & 64& \textbf{79.54 $\pm$ 0.60}\\
Res+KE+LP       & 128   & 83.38 $\pm$  0.18& ViT+KE+LP  & 128 & 78.42 $\pm$ 0.71\\ \bottomrule
\end{tabular}
\label{Assembling results}
\vspace{-1.0cm}
\end{table}

\subsubsection{The Ability to Leverage Class Diversity: }
\begin{figure}[!htb]
    \centering
    \includegraphics[width=\linewidth]{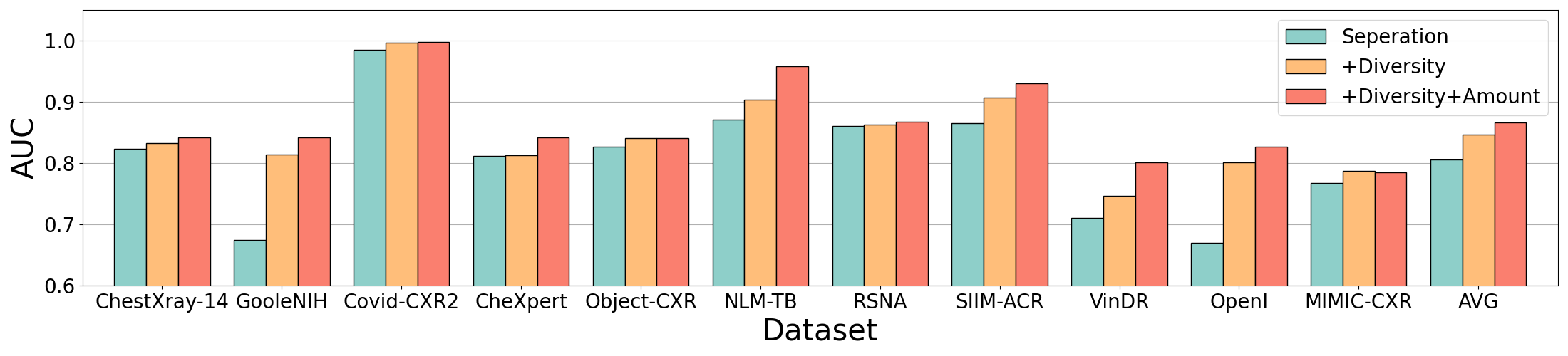}
    \vspace{-0.9cm}
    \caption{Analyse the performance gain on the assembling dataset.  
    ``Seperation'' refers to using a single dataset to train our framework. ``+Diversity'' refers to adding the cases beyond the target classes, increasing the class diversity, and keeping the data amount of the target classes constant. ``+Diversity+Amount'' means directly mixing the 11 datasets and for most datasets, the data amount of the target classes will further increase.
    }
    \label{fig:Results across P}
    \vspace{-0.6cm}
\end{figure}

In this part, we further show that our framework can significantly improve the performance of each dataset, by training on data from other categories. 
Specifically, we consider each dataset separately, \emph{i.e.}, measuring the performance on their own test splits. We propose to decouple the data increment into two dimensions, ``Diversity'' and ``Amount''. ``Diversity'' refers to only adding the cases beyond the target classes and keeping the amount of data of target classes constant, while  ``Amount'' refers to increasing the target class cases. As shown in Fig.~\ref{fig:Results across P}, based on our structure, adding ``Diversity'' can improve the results on all 11 datasets. 
In particular, for some relatively small datasets, the gain is more significant, \emph{e.g.}, GoogleNIH, SIIM-ACR, and OpenI. 
Such an experiment has validated that a knowledge-enhanced model enables to leverage of the shared information between classes, and can be greatly beneficial for dealing with long-tailed diseases, which are seldom annotated in common datasets, 
by leveraging the publicly available data.


\vspace{-0.3cm}

\subsubsection{The Ability to Diagnose Open-set Unseen Diseases: }
\begin{figure}[!htb]
    \centering
    \includegraphics[width=\linewidth]{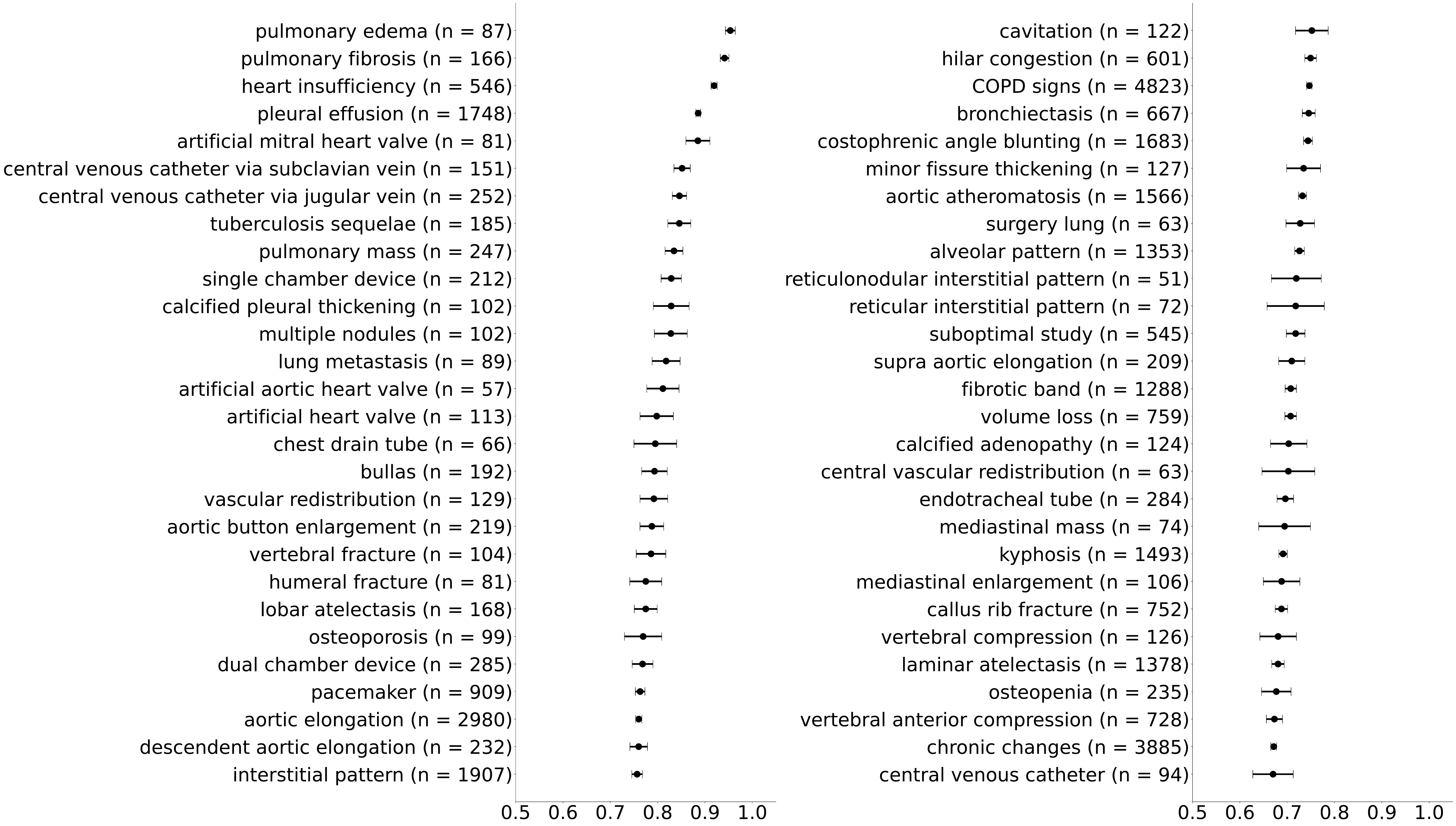}
    \vspace{-.8cm}
    \caption{AUC and $95\%$ CI are shown on the unseen classes under the zero-shot setting. $n$ represents the number of related cases. The top 46 classes are plotted in the figure to show what classes our model can achieve $\text{AUC}>0.700$ on. Generally, our method achieves an AUC of at least 0.800 on 14 findings and at least 0.600 on 79 findings out of 106 radiographic findings where $n>50$ in the PadChest test dataset ($n=39,053$).}
    \label{fig:padchest}
    \vspace{-0.6cm}
\end{figure}

Conventional models can only handle a close-set classification, while, with knowledge-enhanced design, our model enables to predict open-set diseases that never appear in the training set. 
This is meaningful for clinical practical usage, 
as some rare or new diseases can hardly find an off-shelf dataset to re-train models. We test our model on the PadChest testset~\cite{bustos2020padchest} 
and dismiss the classes that exist in the assembling dataset and those having very few cases~($n<=50$), that can hardly have statistically convincing test results. 
During testing, to get the embedding, we simply input the names of unseen classes into the knowledge encoder, and continue the standard evaluation procedure, {\em i.e.}, learnable prompt module, disease-query module. As shown in Fig.~\ref{fig:padchest}, without any example in the training set,  our model can directly achieve an AUC of at least $0.800$ on $14$ findings, at least $0.700$ on $46$ findings and at least $0.600$ on $79$ findings~(unseen at training time) out of 106 radiographic findings. 
This demonstrates our model can break the limits of the labeling classes and be adapted to more practical medical scenarios.

\vspace{-0.5cm}

\section{Conclusion}
\vspace{-.2cm}
In this paper, we propose a novel knowledge-enhanced classification model,  that enables learning visual representation by exploiting the relationship between different medical concepts in the knowledge graph.
While conducting thorough experiments on x-ray image datasets across different anatomy structures, we show that injecting medical prior knowledge is beneficial for tackling (i) long-tailed recognition, (ii) zero-shot recognition.
As for future work, we plan to generalize the idea towards self-supervised learning on pairs of image and text reports and more diverse modalities.



%
%
%
\bibliographystyle{splncs04}
\bibliography{mybib}

\clearpage
\appendix

\section{Detail Results of Knowledge-Enhanced Classification Model}
We show detailed results on the VinDr-PCXR dataset in Tab~\ref{pcxr}, 
VinDr-Mammo dataset in Tab~\ref{mammo}, and VinDr-SpineXr dataset in Tab~\ref{spinexr}.
The proposed knowledge-enhanced classification model exceed the conventional classification model on most diseases.
\textbf{Note that}, our model significantly boosts the performance from $73.29\%$ to $80.07\%$ on Brocho-pneumonia in VinDr-PCXR, demonstrating the benefits of injecting relations between different diseases~(the relation and definition of bronchilits, pneumonia and broncho-pneumonia are shown in the Fig~\ref{fig:intuition}).

\begin{table}[h]
\caption{Compare with Baseline Models on VinDr-PCXR Task.
For each disease, AUC score is reported.
We report the mean and standard deviation of three different seeds.
We use the combine of the first two letters of each word to represent the disease with two words, and first four letters to represent the disease with only one word.
}
\centering
\scriptsize
\setlength{\tabcolsep}{2pt}
\renewcommand{\arraystretch}{1.2}
\begin{tabular}{lp{0.75cm}<{\centering}p{1.2cm}<{\centering}p{1.2cm}<{\centering}p{1.2cm}<{\centering}p{1.2cm}<{\centering}p{1.2cm}<{\centering}p{1.2cm}<{\centering}p{1.2cm}<{\centering}}
\toprule
Model & Prompt & NoFi & Bron & BrPn & OtDi & BrCh & Pneu & Mean \\
 \midrule
ResNet~\cite{he2016deep} & - &  73.88$\pm$1.04 & 67.13$\pm$1.51 & 73.29$\pm$1.76 & 63.52$\pm$1.37 & 68.07$\pm$3.54 & 77.27$\pm$1.24 & 70.53$\pm$1.00 \\
Res+KE & - &  72.85$\pm$1.05 & 67.28$\pm$1.96 & 76.77$\pm$0.55 & 64.51$\pm$0.84 & \textcolor{black}{70.69$\pm$0.46} & 76.28$\pm$2.68 & 71.39$\pm$0.66 \\
Res+KE+LP & 32 & \textbf{75.75$\pm$0.68} & \textbf{71.39$\pm$0.90} & \textcolor{black}{80.01$\pm$1.33} & \textbf{68.15$\pm$1.20} & 69.39$\pm$0.96 & \textbf{79.10$\pm$1.26} & \textbf{73.97$\pm$0.28} \\
Res+KE+LP & 64 & 75.34$\pm$0.20 & 69.91$\pm$1.72 & 78.94$\pm$0.68 & 66.13$\pm$1.31 & 69.91$\pm$1.36 & 77.07$\pm$1.49 & 72.88$\pm$0.19 \\
Res+KE+LP & 128 & \textcolor{black}{75.59$\pm$0.59} &	\textcolor{black}{70.62$\pm$ 0.54} & \textbf{80.07$\pm$0.58} & \textcolor{black}{67.23$\pm$0.32} & \textbf{70.71$\pm$1.88} & \textcolor{black}{77.98$\pm$1.45} &	\textcolor{black}{73.70$\pm$0.41} \\
\midrule
 Train number & & 4629 & 759 & 482 & 442 & 448 & 352 \\
 Test number & & 907 & 174 & 84 & 85 & 90 & 89\\
\bottomrule
\end{tabular}
\label{pcxr}
\end{table}

\begin{table*}[h]
\caption{Compare with Baseline Models on VinDr-Mammo Task.
For each disease, AUC score is reported.
We report the mean and standard deviation of three different seeds.
We use the combine of the first two letters of each word to represent the disease with two words, and first four letters to represent the disease with only one word.}
\centering
\scriptsize
\setlength{\tabcolsep}{1pt}
\renewcommand{\arraystretch}{1.2}
\begin{tabular}{lp{0.75cm}<{\centering}p{0.75cm}<{\centering}p{0.75cm}<{\centering}p{0.75cm}<{\centering}p{0.75cm}<{\centering}p{0.75cm}<{\centering}p{0.75cm}<{\centering}p{0.75cm}<{\centering}p{0.75cm}<{\centering}p{0.75cm}<{\centering}p{0.75cm}<{\centering}p{0.75cm}<{\centering}p{0.75cm}<{\centering}}
\toprule
Model & Prompt & Mass & Asym & NiRe & SkTh & FoAs & SkRe & ArDi & SuCa & SuLN & NoFi & GlAs & Mean\\
 \midrule
ResNet~\cite{he2016deep} & - &  \textbf{78.52 $\pm$ 2.32 } &75.12 $\pm$ 3.49 & \textbf{96.24 $\pm$ 2.52} &\textbf{92.94 $\pm$ 2.70} &75.60 $\pm$ 1.55 &\textbf{98.60 $\pm$ 0.90} &60.35 $\pm$ 3.33 &82.26 $\pm$ 2.13 &97.27 $\pm$  0.30 &76.43 $\pm$ 2.58 &74.59 $\pm$ 6.34 &82.54 $\pm$ 1.79 \\
Res+KE & - & 77.85 $\pm$ 3.80 &71.94 $\pm$ 1.02 &94.33 $\pm$ 3.57 &92.35 $\pm$ 1.50 &73.30 $\pm$ 3.67 &96.31 $\pm$ 2.55 &\textbf{71.97 $\pm$3.36}  &84.30 $\pm$ 1.38 &92.70 $\pm$ 2.74 &77.49 $\pm$ 2.37 &83.05 $\pm$ 6.61  &83.23  $\pm$ 1.98 \\
 Res+KE+LP & 32 & 73.59 $\pm$ 3.82 & 77.38 $\pm$ 2.35 & 90.26 $\pm$ 7.34 & 85.21 $\pm$ 4.49 & 71.74 $\pm$ 4.39 & 91.11 $\pm$13.63 & 67.72 $\pm$ 3.83 & 79.13 $\pm$ 7.55 & 95.76 $\pm$ 2.49 & 74.05 $\pm$3.98  & 80.56 $\pm$ 2.73 & 80.59 $\pm$ 3.11 \\ 
 Res+KE+LP & 64 & 75.16 $\pm$ 2.29 & 75.24 $\pm$ 2.42 & 86.67 $\pm$ 7.37 & 82.38 $\pm$ 5.64 & 72.14 $\pm$ 2.26 & 93.31 $\pm$ 4.61 & 69.81 $\pm$ 1.00 & 82.36 $\pm$ 1.99 & 96.56 $\pm$ 2.08 & 76.31 $\pm$ 1.48 & 84.06 $\pm$ 4.45 & 81.27  $\pm$ 1.83 \\
 Res+KE+LP & 128 & 78.34 $\pm$ 1.32 & \textbf{81.25 $\pm$ 4.54} & 95.01 $\pm$ 3.16 & 91.19 $\pm$ 3.57 & \textbf{75.86 $\pm$ 2.51} & 95.68 $\pm$ 3.56 & 69.71 $\pm$ 7.13 & \textbf{85.02 $\pm$ 1.07} & \textbf{97.14 $\pm$ 1.64} & \textbf{78.50 $\pm$  0.83} & \textbf{85.21 $\pm$ 3.52} & \textbf{84.81  $\pm$ 1.04 }\\
 \hline
 Train number & &  792 & 69 & 25 & 40 & 200 & 13 & 85 & 293 & 38 & 13143 & 17 \\
 Test number &  &219 & 20 & 7 & 12 & 52 & 2 & 24 & 105 & 10 & 3643 & 6\\
 \bottomrule
\end{tabular}
\label{mammo}
\end{table*}

\begin{table}[!htb]
\caption{Compare with Baseline Models on VinDr-SpineXR Task.
For each disease, AUC score is reported.
We report the mean and standard deviation of three different seeds.
We use the combine of the first two letters of each word to represent the disease with two words, and first four letters to represent the disease with only one word.
}
\centering
\scriptsize
\setlength{\tabcolsep}{4pt}
\renewcommand{\arraystretch}{1.2}
\begin{tabular}{lp{0.75cm}<{\centering}p{0.75cm}<{\centering}p{0.75cm}<{\centering}p{0.75cm}<{\centering}p{0.75cm}<{\centering}p{0.75cm}<{\centering}p{0.75cm}<{\centering}p{0.75cm}<{\centering}p{0.75cm}<{\centering}p{0.75cm}<{\centering}}
\toprule
Model~\cite{he2016deep} & Prompt & NoFi & Pste & FoSt & VeCo & OtLe & DSN & Spon & SuIm& Mean \\
\midrule
ResNet & - & 88.96 $\pm$ 0.35 & 90.36 $\pm$ 0.26 & 92.69 $\pm$ 1.30 & 90.89 $\pm$ 0.24 & 71.63 $\pm$ 0.92 & 78.47 $\pm$ 2.93 & 86.62 $\pm$ 0.37 & \textbf{99.20 $\pm$ 0.41} & 87.35 $\pm$ 0.38\\
Res+KE & - & 88.72 $\pm$ 0.56 & 90.48 $\pm$ 0.71 & 90.46 $\pm$ 1.30 & 89.55 $\pm$ 1.30 & \textbf{75.81 $\pm$ 0.93} & 79.08 $\pm$ 0.54 & 88.95 $\pm$ 0.63 & 99.01 $\pm$ 0.09 & 87.76 $\pm$ 0.41 \\
Res+KE+LP & 32 & {89.51 $\pm$ 0.33} & 90.84 $\pm$ 0.34 & 91.92 $\pm$ 0.48 & 91.70 $\pm$ 0.65 & 74.47 $\pm$ 0.91 & 80.59 $\pm$ 0.28 & 89.48 $\pm$ 0.57 & 99.15 $\pm$ 0.35 & 88.46 $\pm$ 0.22 \\
Res+KE+LP  & 64 & \textbf{89.66 $\pm$ 0.23} & \textbf{90.92 $\pm$ 0.21} & \textbf{92.65 $\pm$ 1.33} & \textbf{92.61 $\pm$ 0.82} & 74.26 $\pm$ 0.46 & \textbf{82.65 $\pm$ 0.93} & 89.61 $\pm$ 0.99 & 98.81 $\pm$ 0.28 & \textbf{88.90 $\pm$ 0.04 }\\
Res+KE+LP  & 128 & 89.35 $\pm$ 0.29 & 90.81 $\pm$0.10  & 91.67 $\pm$ 2.23 & 92.00 $\pm$ 1.39 & 72.41 $\pm$ 1.55 & 80.55 $\pm$ 0.59& \textbf{89.82 $\pm$2.54} & 99.16 $\pm$ 0.36 & 88.22 $\pm$ 0.53 \\
\midrule
Train number & &  4260 & 3575 & 271 & 157 & 333 & 602 & 257 & 257 \\
Test number & & 1070 & 878 & 60 & 52 & 80 & 148 & 62 & 64 \\
\bottomrule
\end{tabular}
\label{spinexr}
\end{table}

\end{document}